\documentclass[10pt]{article}

\usepackage[letterpaper,margin=1in]{geometry}
\usepackage{times}
\usepackage{natbib}
\setcitestyle{authoryear,round,citesep={;},aysep={,},yysep={;}}

\usepackage{graphicx}
\usepackage{booktabs}
\usepackage{multirow}
\usepackage{amsmath}
\usepackage{amssymb}
\usepackage{enumitem}
\usepackage{capt-of}
\usepackage{algorithm}
\usepackage{algpseudocode}
\usepackage{tikz}
\usepackage{url}
\usepackage{hyperref}

\usepackage{amsmath,amsfonts,bm}

\def\eqref#1{equation~\ref{#1}}

\def\1{\bm{1}}

\DeclareMathAlphabet{\mathsfit}{\encodingdefault}{\sfdefault}{m}{sl}
\SetMathAlphabet{\mathsfit}{bold}{\encodingdefault}{\sfdefault}{bx}{n}

\graphicspath{{figures/}}

\title{ECHOES: Seeing Beyond the Current Image via Recursive History Compression and Privileged On-Policy Distillation}

\author{%
  {\normalsize\bfseries
   Tengda Guo$^{1}$ \quad Jiawang Hu$^{2}$ \quad Yifei Tao$^{3}$ \quad Yin Wang$^{2}$\\[0.2em]
   Haotian Hu$^{4}$ \quad Fanyi Wang$^{5}$ \quad Liujuan Tang$^{2}$}
  \\[0.65em]
  {\small
    $^{1}$Tsinghua University \quad
    $^{2}$StepFun (Shanghai Jieyue Xingchen Intelligent Technology Co., Ltd.)\\[0.2em]
    $^{3}$Nanyang Technological University \quad
    $^{4}$Zhejiang Leapmotor Technology Co., Ltd. \quad
    $^{5}$Honor
  }
}
\date{}

\makeatletter
\renewcommand{\@maketitle}{%
  \newpage
  \null
  \vspace*{-2.4em}%
  \noindent\rule{\textwidth}{4pt}\par
  \vspace{0.9em}
  {\centering\LARGE\bfseries\@title\par}
  \vspace{0.9em}
  \noindent\rule{\textwidth}{0.6pt}\par
  \vspace{0.8em}
  {\centering\@author\par}
  \vspace{1.1em}
}
\makeatother

\hypersetup{
  pdftitle={ECHOES: Seeing Beyond the Current Image via Recursive History Compression and Privileged On-Policy Distillation},
  pdfauthor={Tengda Guo, Jiawang Hu, Yifei Tao, wangyin, Haotian Hu, Fanyi Wang, Liujuan Tang}
}

\begin{document}

\maketitle

\begin{abstract}
Long-horizon GUI agents rely on visual history to maintain task states, but retaining historical screenshots increases visual-token usage and inference costs, while textual summaries can discard critical state information, undermining long-horizon reliability. We propose ECHOES, which leverages training-time privileged information to guide textual history compression and strengthen long-horizon reasoning. Specifically, ECHOES adopts a Compression–Think–Action (CTA) format to recursively update a textual history state and reason. The same model, given privileged multi-image inputs, serves as a teacher that provides dense token-level on-policy distillation on student-generated CTA rollouts. Our Perception–Memory Fidelity (PMF) Gap-guided OPD–RL framework dynamically balances privileged distillation and task-reward optimization, with periodic teacher updates supporting continued improvement. We further introduce ECHOESBMK, a fine-grained diagnostic benchmark derived from real GUI agent failure trajectories to jointly evaluate visual perception and cross-modal state retention. Experiments on four GUI benchmarks and ECHOESBMK demonstrate substantial gains in task performance and memory fidelity. Without retaining historical screenshots at deployment, ECHOES achieves competitive performance with a favorable performance–inference-cost trade-off and exhibits positive zero-shot transfer to unseen visual planning tasks.

\end{abstract}

\section{Introduction}
\label{sec:introduction}

Recent advances in multimodal large language models (MLLMs) have enabled agent systems capable of perceiving, reasoning, and acting in complex visual environments, with increasing success in long-horizon visual reasoning tasks~\citep{xu2025aguvis,liu2025guirise}. As a representative application, graphical user interface (GUI) agents have made substantial progress in single-step visual decision-making and grounding~\citep{cheng2024seeclick,lin2025showui}, yet remain brittle in long-horizon control tasks~\citep{tian2025agentprog,zhou2026hiconagent}. A key cause is state-retention failure, which causes decisions and execution to drift out of alignment with the actual environment state. As interaction trajectories grow, long-horizon GUI agents must therefore not only understand the current interface but also preserve past states that remain relevant to future decisions. Existing approaches support long-horizon reasoning by retaining visual history~\citep{zhou2026hiconagent}, generating textual summaries~\citep{liu2025guirise}, or maintaining structured states~\citep{tian2025agentprog}. However, these gains exhibit diminishing returns as trajectories lengthen, while accumulating visual tokens, context length, and KV-cache overhead hinder efficient deployment~\citep{huang2025guikv,zhou2026hiconagent}. Long-horizon visual agents thus face a fundamental tension: additional historical screenshots increase inference costs, whereas directly converting visual history into text can omit critical task states through cross-modal information loss. We term this loss the {Perception--Memory Fidelity Gap (PMF Gap)}. This motivates a different question: {Can multi-image visual history be used only during training to transfer its decision-making capabilities to a model that relies solely on the current image and low-cost reasoning over compressed history at deployment?}

\begin{figure}[t]
    \centering
    \includegraphics[width=\linewidth]{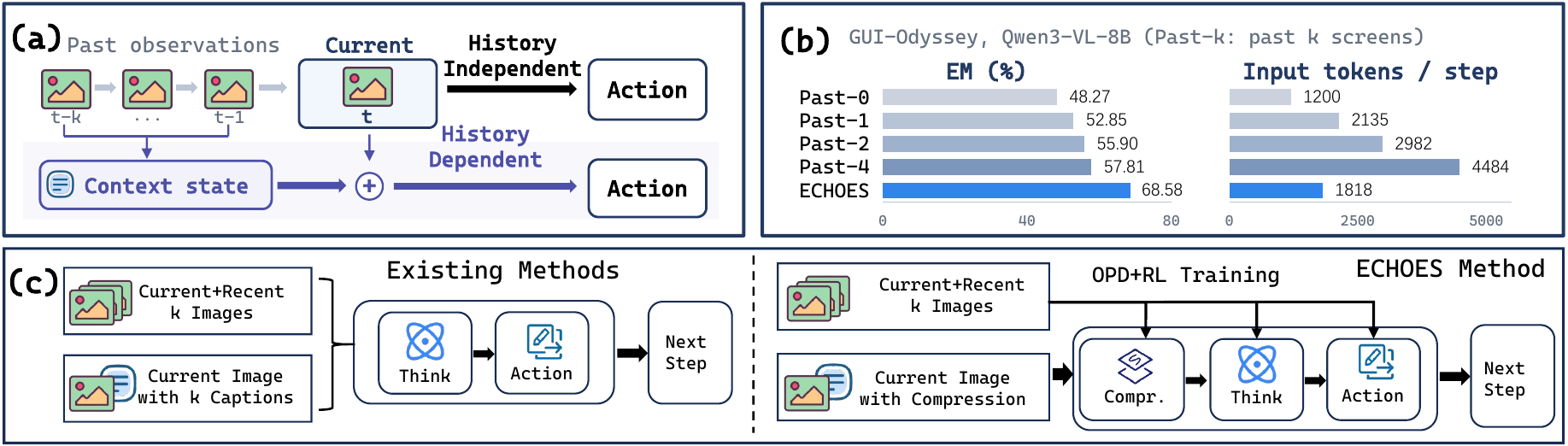}
    \vspace{-10px}
    \caption{\textbf{Motivation and method comparison.} (a) History-independent versus history-dependent decisions. (b) ECHOES's performance--cost advantage on GUI-Odyssey. (c) Existing methods versus ECHOES with recursive CTA and training-time multi-image supervision.}
    \label{fig:intro}
    \vspace{-15px}
\end{figure}

To address this question, we propose {ECHOES}, a training method for long-horizon visual agents, as illustrated in Figure~\ref{fig:intro}. We model task-relevant compressed history as a {learnable approximate sufficient statistic} of multi-image visual history for subsequent decisions. We first adopt the {Compression--Think--Action (CTA)} reasoning mode, which continuously maintains long-horizon task states through recursive Compression. We then introduce training-time privileged on-policy distillation (OPD)~\citep{agarwal2024onpolicy}, using the stronger multi-image \emph{privileged view} as the teacher to provide dense token-level supervision on CTA trajectories generated by the single-image + Compression \emph{deployable view} itself. This encourages the textual state to progressively preserve information in the full visual history that is relevant to subsequent decisions. Building on this design, we construct a {PMF Gap-guided OPD--RL} training framework that dynamically coordinates the optimization weights of OPD and GRPO~\citep{shao2024deepseekmath} across rounds according to the information gap between the privileged and deployable observation modes. When the gap is large, the framework emphasizes privileged-policy supervision; as it narrows, task-reward optimization receives greater weight, gradually shifting learning from absorbing the privileged policy's information advantage to directly optimizing the actual task objective. To evaluate GUI agents' long-horizon performance at a finer granularity, we additionally introduce {ECHOESBMK}, a benchmark constructed from real failure trajectories that spans seven task categories across three levels and jointly evaluates perception and cross-modal information retention.

Experiments show that ECHOES substantially improves performance on long-horizon GUI benchmarks and ECHOESBMK with lower inference overhead, while demonstrating strong zero-shot generalization to non-GUI visual planning environments. Further experiments reveal pronounced improvements on longer-horizon and history-dependent tasks, and show that these gains stem from the CTA reasoning mode and the training design. In summary, our main contributions are as follows:

\begin{enumerate}[labelindent=0pt, leftmargin=*, itemsep=2pt, topsep=2pt]
    \item We propose {ECHOES}, a training-time privileged OPD method that adopts CTA reasoning and uses additional historical visual information to provide dense supervision, transferring the decision-making advantage of multi-image inputs to a low-cost deployment mode that relies only on the current observation and a recursive textual state.

    \item We propose a {PMF Gap-guided OPD--RL} training framework that dynamically balances privileged distillation and task-reward optimization according to the privileged--deployable information gap. A {round-wise self-evolving teacher} continuously strengthens both observation modes, substantially improving long-horizon visual decision-making.

    \item We construct {ECHOESBMK}, a fine-grained benchmark derived from real GUI agent failure trajectories, to jointly evaluate visual perception and memory fidelity in the conversion of visual information into textual states. We further demonstrate the advantages of ECHOES across multiple benchmarks and analyze its performance along multiple dimensions.
\end{enumerate}

\section{Related Work}
\label{sec:related}

Long-horizon VLM agents require both effective history representations and policies trained for sequential decision-making. Beyond visual grounding and action modeling~\citep{cheng2024seeclick,qin2025uitars}, recent approaches use textual summaries, history-aware policy optimization, and latent memory to preserve decision-relevant context~\citep{liu2025guirise,zhou2026hiconagent,zhang2026memw}. These approaches address the trade-off between historical information and inference cost, including learning to align compressed and richer context representations. In parallel, reinforcement learning improves GUI policies through task rewards~\citep{lu2026uir1,luo2025guir1}, while on-policy distillation supplies teacher feedback on student-generated sequences~\citep{agarwal2024onpolicy}. Recent studies further explore target-region cues and future observations as privileged contexts~\citep{zhang2026guisd,li2026ghd}, as well as combining distillation with reinforcement learning~\citep{wu2026litegui}. Building on these directions, our work targets the capability gap between multi-image historical observations and single-image deployment with recursive textual compression. We use recent visual history as training-time privileged evidence to supervise complete Compression--Think--Action sequences, jointly guiding state updates, reasoning, and action generation. With a round-wise refreshed teacher, we adapt the distillation--RL balance to the measured privileged--deployable performance gap. This design transfers history-informed decision-making into the deployable policy without retaining historical screenshots at inference. Appendix~\ref{app:total_related_work} provides an extended discussion.

\vspace{-10px}
\section{Method}
\vspace{-10px}
\label{sec:method}

\subsection{Compression--Think--Action Reasoning and Task-State Modeling}

\begin{figure}[t]
    \centering
    \includegraphics[width=\linewidth]{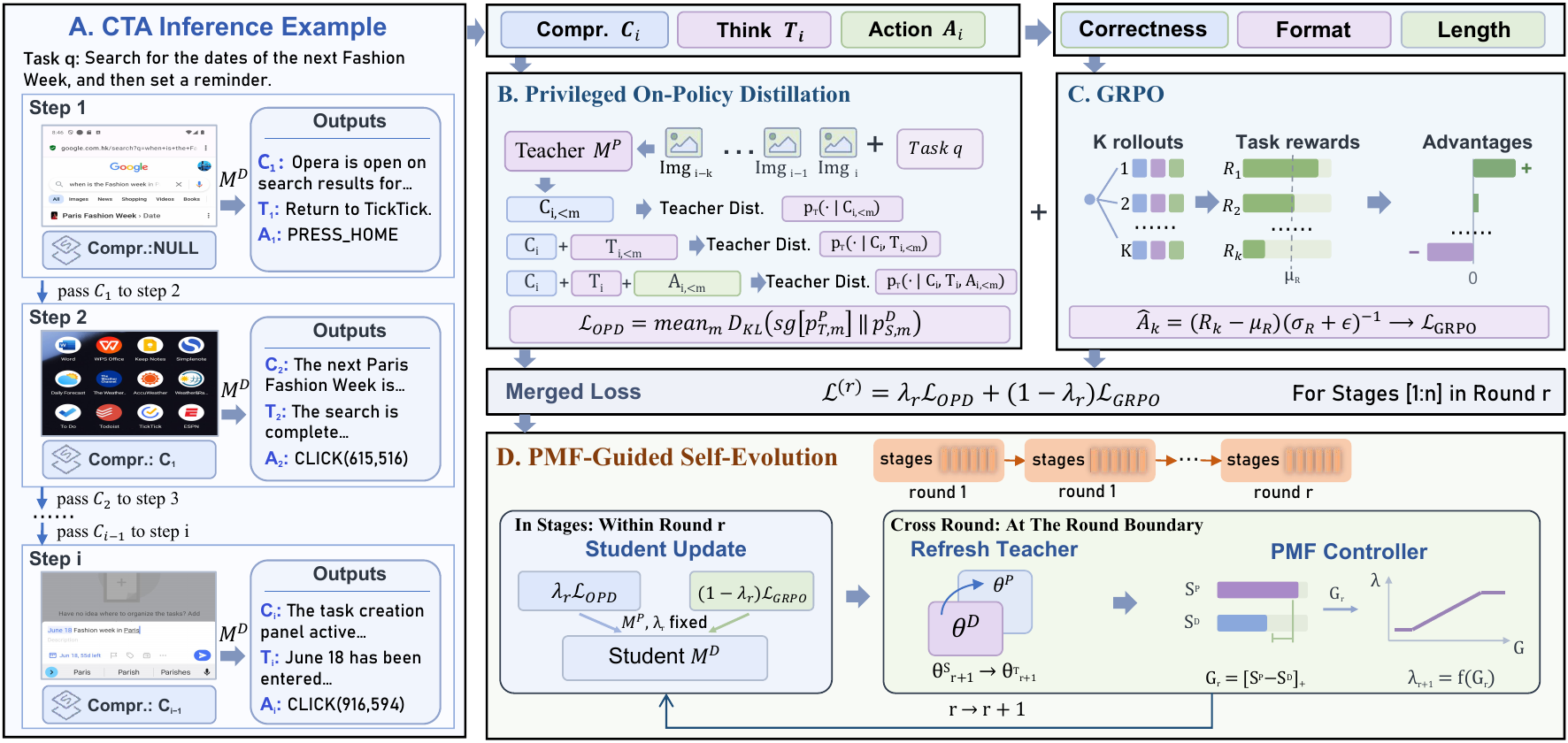}
    \vspace{-20px}
    \caption{\textbf{Overview of ECHOES.} Given the task, current screenshot, and previous $C$, the policy generates a CTA sequence. Training combines OPD from the same model's privileged multi-image view with GRPO under PMF-guided weighting, refreshing the teacher at round boundaries.}
    \label{fig:framework}
\end{figure}

Long-horizon visual decision-making requires historical context beyond the current observation, yet accumulating screenshots increases visual-token usage and inference cost with trajectory length. Since relevant history varies across decision stages, the model should autonomously select, compress, and reuse historical information rather than rely on a fixed external summarizer. We therefore adopt {Compression--Think--Action (CTA)}, which integrates recursive textual-state compression, explicit reasoning, and action generation into a single autoregressive sequence, as shown in Figure~\ref{fig:framework}.

At step $t$, the deployment state comprises the task instruction $q$, current visual observation $o_t$, and previous compression $C_{t-1}$. Conditioned on this state, the model generates the complete CTA sequence in a single autoregressive rollout:
\[
s_t=(q,o_t,C_{t-1}), \qquad (C_t,T_t,A_t)\sim\pi_\theta(\cdot\mid s_t).
\]
Here, $C_t$ is the updated textual task state, $T_t$ is explicit reasoning, and $A_t$ belongs to a unified nine-category GUI action space. After executing $A_t$, the environment returns $o_{t+1}$, while only $C_t$ is retained across steps. CTA thus maintains historical context by \textbf{recursively rewriting its textual state} instead of accumulating screenshots.

Compression aims to preserve decision-relevant information, not reconstruct the past. Let $H_t^V$ denote the complete visual interaction history available at step $t$. Ideally, the compressed deployment state should support a decision distribution approaching that conditioned on the full history:
\[
D_{\mathrm{KL}}\!\left(
p(T_t,A_t\mid q,H_t^V)
\,\big\|\,
p(T_t,A_t\mid s_t)
\right)
\longrightarrow 0,
\qquad |C_t|\le B,
\]
where $B$ denotes the textual capacity budget and the second distribution marginalizes over the generated $C_t$. This design aims to learn how to encode and recursively update historical context in $C_t$, such that the resulting joint distribution over reasoning and actions approximates its full-history counterpart. Under this budget, $C_t$ should prioritize task progress, critical state changes, completed operations, and observed execution outcomes, while discarding details irrelevant to future decisions.

\subsection{Multi-Image--Single-Image Training-Time Privileged Self-Distillation}

\setlength{\textfloatsep}{6pt plus 1pt minus 2pt}
\begin{algorithm}[t]
\caption{PMF Gap-Guided Privileged OPD--RL}
\label{alg:pmf_opd_rl}
\begin{algorithmic}[1]
\Require Initial student $\theta$, fixed validation set $\mathcal{D}_{\mathrm{val}}$,
rounds $R$, weight bounds $\lambda_{\min},\lambda_{\max}$, $\epsilon>0$
\Ensure Trained deployable CTA policy $\pi_\theta^D$
\State $G_0 \gets \max\{0,S^P(\theta)-S^D(\theta)\}$
\State Define $f(g)=\operatorname{clip}\!\left(
\lambda_{\min}+(\lambda_{\max}-\lambda_{\min})
\frac{g}{G_0+\epsilon},
\lambda_{\min},\lambda_{\max}\right)$;
$\lambda_1 \gets f(G_0)$
\For{$r=1,\ldots,R$}
    \State $\bar{\theta} \gets \operatorname{copy}(\theta)$
    \Comment{Freeze teacher within round}
    \For{each student update in round $r$}
        \State Sample grouped CTA rollouts
        $Y_t=(C_t,T_t,A_t)\sim
        \pi_\theta^D(\cdot\mid q,o_t,C_{t-1})$
        \State $p_{t,i}^{P}\gets
        \pi_{\bar{\theta}}(\cdot\mid q,o_t,\mathcal{V}_t,Y_{t,<i}),
        \quad
        p_{t,i}^{D}\gets
        \pi_{\theta}(\cdot\mid q,o_t,C_{t-1},Y_{t,<i})$
        \State $\mathcal{L}_{\mathrm{OPD}}^{\mathrm{CTA}}
        \gets \operatorname{mean}_{\text{valid CTA tokens}}
        D_{\mathrm{KL}}\!\left(
        \operatorname{sg}[p_{t,i}^{P}]
        \,\|\,p_{t,i}^{D}\right)$
        \State $\theta\gets\operatorname{Update}\!\left(
        \theta,\,
        \lambda_r\mathcal{L}_{\mathrm{OPD}}^{\mathrm{CTA}}
        +(1-\lambda_r)\mathcal{L}_{\mathrm{GRPO}}^{\mathrm{CTA}}
        \right)$
    \EndFor
    \State $G_r\gets\max\{0,S^P(\theta)-S^D(\theta)\},
    \quad \lambda_{r+1}\gets f(G_r)$
\EndFor
\end{algorithmic}
\end{algorithm}

CTA requires compression to preserve historical information needed for decisions. However, supervision confined to the compressed view does not expose decision-relevant visual evidence omitted from textual state. We therefore introduce {training-time privileged self-distillation} based on on-policy distillation (OPD)~\citep{agarwal2024onpolicy}, using the same model's multi-image view as teacher to guide the deployable single-image CTA policy. This observation asymmetry makes visual history a training-time reference for learning recursive compression and history-informed decisions, rather than a deployment input. At step $t$, the student and teacher receive:

\vspace{-5pt}
\[
x_t^{D}=(q,o_t,C_{t-1}), \qquad
x_t^{P}=(q,o_t,\mathcal{V}_t),
\]
where D and P denote deployable and privileged views, and $V_t$ contains recent historical screenshots preceding step $t$. The teacher uses the same architecture as the student, exploiting additional visual evidence rather than greater model capacity. The student samples $Y_t=(C_t,T_t,A_t)\sim\pi_{\theta}^{D}(\cdot\mid x_t^{D})$ under deployment inputs. Both policies are then evaluated on the same student-generated prefixes via teacher-forced scoring:
\[
p_{t,i}^{P}
=
\pi_{\bar{\theta}}
\left(
\cdot\mid x_t^{P},Y_{t,<i}
\right),
\qquad
p_{t,i}^{D}
=
\pi_{\theta}
\left(
\cdot\mid x_t^{D},Y_{t,<i}
\right),
\]
where $\bar{\theta}$ denotes the teacher parameters and $Y_{t,<i}$ the student's token prefix. We distill all Compression, Think, and Action tokens using
\[
\mathcal{L}_{\mathrm{OPD}}^{\mathrm{CTA}}
=
\frac{1}{N}
\sum_{i\in
\mathcal{I}_{C}\cup\mathcal{I}_{T}\cup\mathcal{I}_{A}}
D_{\mathrm{KL}}
\left(
\operatorname{sg}[p_{t,i}^{P}]
\,\|\,p_{t,i}^{D}
\right),
\]
where $\mathcal{I}_{C}$, $\mathcal{I}_{T}$, and $\mathcal{I}_{A}$ denote the corresponding token positions, $N$ counts valid distilled tokens, and $\operatorname{sg}[\cdot]$ stops teacher gradients. Supervising $C_t$ guides the retention and recursive updating of historical information, while supervising $T_t$ and $A_t$ aligns reasoning and action generation with the teacher's richer visual context. Together, these signals train the deployable CTA policy to approximate its privileged counterpart:
\[
\pi_{\theta}^{D}(\cdot\mid q,o_t,C_{t-1})
\;\xrightarrow{\;\mathrm{OPD}\;}\;
\pi_{\bar{\theta}}^{P}(\cdot\mid q,o_t,\mathcal{V}_t).
\]
The privileged visual history is used only during training; inference retains the single-image + Compression input without additional visual or teacher overhead.

\vspace{-5pt}
\subsection{PMF Gap-Guided OPD--RL Training Framework}
\vspace{-5pt}

OPD provides dense supervision but targets privileged-policy behavior, whereas GRPO directly optimizes task rewards from relatively sparse feedback. We introduce {PMF Gap-guided OPD--RL} to balance knowledge transfer and reward-driven improvement according to the remaining advantage of privileged visual history. Rather than using a fixed mixture, this framework adapts the balance to the model's evolving performance under the two observation modes. Each round comprises deployable CTA rollouts and student updates using OPD (Sec.~3.2) and GRPO~\citep{shao2024deepseekmath}:
\[
\mathcal{L}^{(r)}
=
\lambda_r \mathcal{L}_{\mathrm{OPD}}^{\mathrm{CTA}}
+
(1-\lambda_r)\mathcal{L}_{\mathrm{GRPO}}^{\mathrm{CTA}}.
\]
Both losses are averaged over their respective valid tokens or rollout samples before combination. GRPO rewards action correctness, valid and parseable CTA formatting, and penalizes excessive Compression length. We adopt a {round-wise self-evolving teacher}, with student and teacher parameters $\theta_r$ and $\bar{\theta}_r$. The teacher and $\lambda_r$ remain fixed within each round; the updated student refreshes the teacher at the round boundary:
\[
\bar{\theta}_{r+1}\leftarrow\theta_{r+1},
\qquad
\nabla_{\bar{\theta}_r}\mathcal{L}^{(r)}=0.
\]

At each round's end, the latest student is evaluated under both views on a fixed small-scale validation set, yielding scores $S_r^P$ and $S_r^D$. Their nonnegative difference defines the training-time {PMF Gap}:
\[
G_r=\max(0,S_r^P-S_r^D).
\]
Because both scores use the same model parameters, this gap reflects the relative benefit of privileged visual history rather than differences between model checkpoints. Let $G_0$ denote the initial gap before training. We map the current gap relative to $G_0$ to the next-round OPD weight:
\[
\lambda_{r+1}
=
\operatorname{clip}
\left(
\lambda_{\min}
+
(\lambda_{\max}-\lambda_{\min})
\frac{G_r}{G_0+\epsilon},
\lambda_{\min},
\lambda_{\max}
\right).
\]
Here, $\epsilon>0$ ensures numerical stability, and $[\lambda_{\min},\lambda_{\max}]$ bounds the weight. Larger gaps favor OPD, whereas shrinking gaps shift weight toward GRPO. The weight can increase again if the gap widens, enabling gap-responsive adaptation rather than a prescribed decay. This balances privileged-policy transfer with reward-driven improvement beyond imitation.

\vspace{-10pt}
\section{Benchmark}
\label{sec:benchmark}
\vspace{-5pt}
\subsection{Failure-Driven GUI Perception and Memory-Fidelity Evaluation}
\vspace{-5pt}
Existing GUI benchmarks cover grounding, screen understanding, and task execution~\citep{cheng2024seeclick,hsiao2025screenqa,rawles2023android,xie2024osworld,lu2025guiodyssey}, but overall success rates conflate perception, memory, planning, and action errors, obscuring the bottlenecks of long-horizon agents. Meanwhile, fine-grained evaluations primarily assess information acquisition from visual inputs~\citep{cheng2024seeclick,hsiao2025screenqa}, with less attention to whether this information remains usable after compression into textual memory. Our failure analysis shows that click and filtering-state errors are among the most common failure modes, suggesting that screen perception and state retention may underlie many execution errors. We therefore construct a {failure-driven diagnostic benchmark} that jointly examines {visual state understanding and its preservation in textual memory} for fine-grained analysis of long-horizon execution. Specifically, we evaluate {Perception} together with the {Perception--Memory Fidelity Gap (PMF Gap)}, defined as $G_{\mathrm{PMF}}=\max(0,S^V-S^M)$, where $S^V$ and $S^M$ denote normalized scores using direct visual input and only model-generated textual memory, respectively. A smaller gap indicates less performance degradation after compression; interpreted alongside direct-vision performance, it helps distinguish failures to understand visual states from failures to retain the acquired state information.

\begin{figure}[t]
\centering
\begin{tikzpicture}[
    x=\linewidth, y=12pt,
    font=\small,
    inner sep=0pt,
    outer sep=0pt
]
    \node[anchor=south west] at (0.58,-1.05) {
        \includegraphics[
            width=0.39\linewidth,
            height=98.4pt
        ]{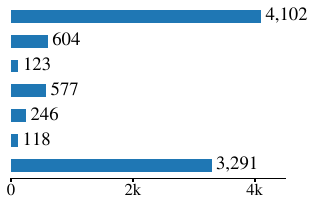}
    };

    \node[anchor=west,font=\small\bfseries]
        at (0,7.28) {Level};
    \node[anchor=west,font=\small\bfseries]
        at (0.09,7.28) {Task};
    \node[anchor=west,font=\small\bfseries]
        at (0.338,7.28) {Diagnostic target};
    \node[anchor=west,font=\small\bfseries]
        at (0.594,7.28) {\# Instances};

    \draw[line width=0.5pt]
        (0,6.96) -- (1,6.96);

    \node[anchor=west,font=\small\bfseries]
        at (0,6) {L1};
    \node[anchor=west,font=\small\bfseries]
        at (0,3) {L2};
    \node[anchor=west,font=\small\bfseries]
        at (0,0.5) {L3};

    \foreach \yy/\task/\target in {
        6.5/{BBox-to-Function}/{Element semantics},
        5.5/{Semantic-to-BBox}/{Visual grounding},
        4.5/{Active-Tab}/{Navigation state},
        3.5/{Foreground-Layer}/{Foreground state},
        2.5/{Blocking-State}/{Interaction blocking},
        1.5/{Screen-Selection}/{Selection state},
        0.5/{Screen-State Diff}/{Cross-frame changes}
    } {
        \node[anchor=west]
            at (0.09,\yy) {\task};
        \node[anchor=west]
            at (0.338,\yy) {\target};
    }

    \draw[line width=0.25pt]
        (0,5) -- (1,5);
    \draw[line width=0.25pt]
        (0,1) -- (1,1);

    \draw[line width=0.5pt]
        (0,0) -- (1,0);

\end{tikzpicture}
\vspace{-20pt}
\caption{\textbf{Task taxonomy and statistics of ECHOESBMK.} Seven tasks span element-level perception (L1), whole-screen state understanding (L2), and cross-frame state changes (L3). Bars indicate evaluation-instance counts.}
\label{fig:benchmark_overview}
\vspace{0pt}
\end{figure}

Following the observed failure patterns, we derive seven diagnostic tasks from real GUI agent rollouts and failure records by unifying screenshots, component trees, and trajectories into standardized screen states. These tasks span {L1 element-level perception}, {L2 whole-screen state understanding}, and {L3 cross-frame state changes}. Single-screen tasks use interface elements and state annotations, while cross-frame tasks sample pre- and post-action observations to preserve state transitions from real execution. Tasks retain their natural output formats, including coordinate localization, categorical judgments, and natural-language descriptions, rather than being uniformly converted into action prediction. This reduces confounding from policy behavior and action-space design when measuring underlying perception capabilities.
\vspace{-10pt}
\subsection{Benchmark Statistics and Quality Control}

Figure~\ref{fig:benchmark_overview} summarizes the task taxonomy, diagnostic targets, and instance distributions. The benchmark combines dense local queries with whole-screen and cross-frame evaluations. Automatic construction and LLM-assisted annotation are followed by {manual verification} to remove ambiguous or incorrect samples, and evaluation uses metrics matched to each task's output. Construction details, counting conventions, annotation checks, and evaluation protocols are provided in Appendix~\ref{app:benchmark}.

\vspace{-5pt}
\section{Experiments}
\vspace{-5pt}
\label{sec:experiments}

\subsection{Experimental Setup}
\vspace{-5pt}

\noindent\textbf{Training.}
We first perform CTA SFT on {Qwen3-VL-8B}~\citep{bai2025qwen3vl} using {10K} self-constructed samples, followed by privileged OPD--RL on {9.5K} samples with PMF Gap-guided weighting.

\noindent\textbf{Benchmarks and baselines.}
We evaluate on {GUI-Odyssey}~\citep{lu2025guiodyssey}, {AndroidControl-v2-High}~\citep{li2024datascale}, {AITW}~\citep{rawles2023android}, and {AMEX-L3}~\citep{chai2025amex}, with {MageBench-Sokoban}~\citep{zhang2026magebench} for zero-shot transfer and {ECHOESBMK} (Sec.~4) for perception and memory-fidelity diagnostics.
Baselines include general VLMs from the Qwen, GLM, and Step3 families~\citep{bai2025qwen3vl, bai2025qwen25vl, qwen2026qwen35, glm2025vl, huang2026step3vl} (GLM-4.1V-9B denotes the Thinking variant); GUI agents MAI-UI-8B~\citep{zhou2025maiui}, AgentCPM-GUI-8B~\citep{zhang2025agentcpm}, GUI-Owl-1.5-8B/32B~\citep{xu2026mobileagent35}, UI-TARS-1.5-7B~\citep{qin2025uitars}, UI-R1-3B~\citep{lu2026uir1}, and InfiGUI-R1-3B~\citep{liu2025infigui}; and proprietary models GPT-5.5~\citep{openai2026gpt55} and Gemini 3.1 Pro~\citep{googledeepmind2026gemini31}.

\noindent\textbf{Evaluation protocol.}
Baselines use favorable model-specific protocols, following official/native inference procedures where available and retaining native or GT history where applicable.
Unless otherwise specified, ECHOES receives only the task instruction, current screenshot, and self-generated recursive $C$, without historical screenshots or GT history.
Efficiency is measured by per-step input token counts and wall-clock latency.
Evaluation and training details appear in Appendices~\ref{app:experimental_details} and~\ref{app:training_details}, respectively.

\vspace{-5pt}
\subsection{Main Results, Efficiency, and Transfer}

\begin{table}[t]
\centering
\caption{Main results on four GUI benchmarks. All scores are reported in percentage (\%). Best and second-best results are marked in \textbf{bold} and \underline{underline}, respectively.}
\label{tab:main_results}
\setlength{\tabcolsep}{2.6pt}
\renewcommand{\arraystretch}{1.08}
\begin{tabular}{l cc ccc cc cc}
\toprule
\multirow{2}{*}{Model}
& \multicolumn{2}{c}{GUI-Odyssey}
& \multicolumn{3}{c}{AC-v2-High}
& \multicolumn{2}{c}{AITW}
& \multicolumn{2}{c}{AMEX-L3} \\
\cmidrule(lr){2-3}\cmidrule(lr){4-6}\cmidrule(lr){7-8}\cmidrule(lr){9-10}
& TM $\uparrow$ & EM $\uparrow$
& Acc $\uparrow$ & Type $\uparrow$ & Ground $\uparrow$
& Step $\uparrow$ & Type $\uparrow$
& Step $\uparrow$ & Type $\uparrow$ \\
\midrule

\multicolumn{10}{c}{\textit{Proprietary / Closed-source Models}} \\
\midrule
GPT-5.5             & 73.84 & 48.76 & 46.92 & 69.85 & 64.38 & 48.65 & 63.72 & 68.84 & 80.92 \\
Gemini 3.1 Pro      & 77.18 & 54.92 & 50.82 & 74.36 & \underline{68.22} & 53.18 & 65.46 & 72.36 & 83.46 \\

\midrule
\multicolumn{10}{c}{\textit{Open-source Models}} \\
\midrule
Qwen3-VL-8B         & 76.14 & 48.27 & 49.62 & 63.28 & 64.43 & 44.16 & 56.40 & 62.28 & 80.10 \\
UI-R1-3B            & 59.30 & 38.40 & 43.20 & 58.10 & 55.60 & 52.30 & 64.10 & 60.70 & 75.80 \\
InfiGUI-R1-3B       & 77.86 & 56.72 & 50.61 & 75.84 & 67.92 & 56.48 & 65.62 & 71.84 & 82.72 \\
Step3-VL-10B        & 77.42 & 50.86 & 50.08 & 68.74 & 66.58 & 46.92 & 60.18 & 65.73 & 80.84 \\
MAI-UI-8B           & \underline{81.34} & \underline{65.82} & 51.68 & 76.12 & \textbf{68.76} & 56.72 & {62.88} & \underline{73.18} & \underline{84.02} \\
GLM-4.1V-9B         & 75.33 & 48.90 & 39.20 & 67.33 & 51.32 & 16.00 & 48.00 & 21.33 & 41.30 \\
Qwen3-VL-32B        & 79.79 & 54.35 & \underline{52.53} & 75.13 & 63.34 & 47.36 & 61.00 & 71.61 & 83.00 \\
AgentCPM-GUI-8B     & 80.53 & 64.61 & 43.37 & \underline{76.84} & 59.44 & \underline{57.09} & \textbf{71.20} & 69.54 & 82.40 \\
GUI-Owl-1.5-8B      & 79.24 & 58.95 & 41.12 & 59.75 & 60.81 & 51.80 & 63.50 & 72.48 & 81.70 \\
UI-TARS-1.5-7B      & 74.53 & 52.03 & 45.20 & 70.60 & 57.40 & 34.70 & 51.80 & 42.31 & 57.50 \\

\midrule
\multicolumn{10}{c}{\textit{Ours / Same-backbone Comparison}} \\
\midrule
\textbf{Ours-8B}    & \textbf{82.25} & \textbf{68.58} & \textbf{53.60} & \textbf{76.96} & 65.56 & \textbf{59.35} & \underline{66.30} & \textbf{73.90} & \textbf{84.60} \\
$\Delta$ over Base  & +6.11 & +20.31 & +3.98 & +13.68 & +1.13 & +15.19 & +9.90 & +11.62 & +4.50 \\

\bottomrule
\end{tabular}
\end{table}

\begin{figure}[t]
\centering

\begin{minipage}[t]{0.69\linewidth}
    \centering
    \vspace{0pt}

    \includegraphics[width=\linewidth]{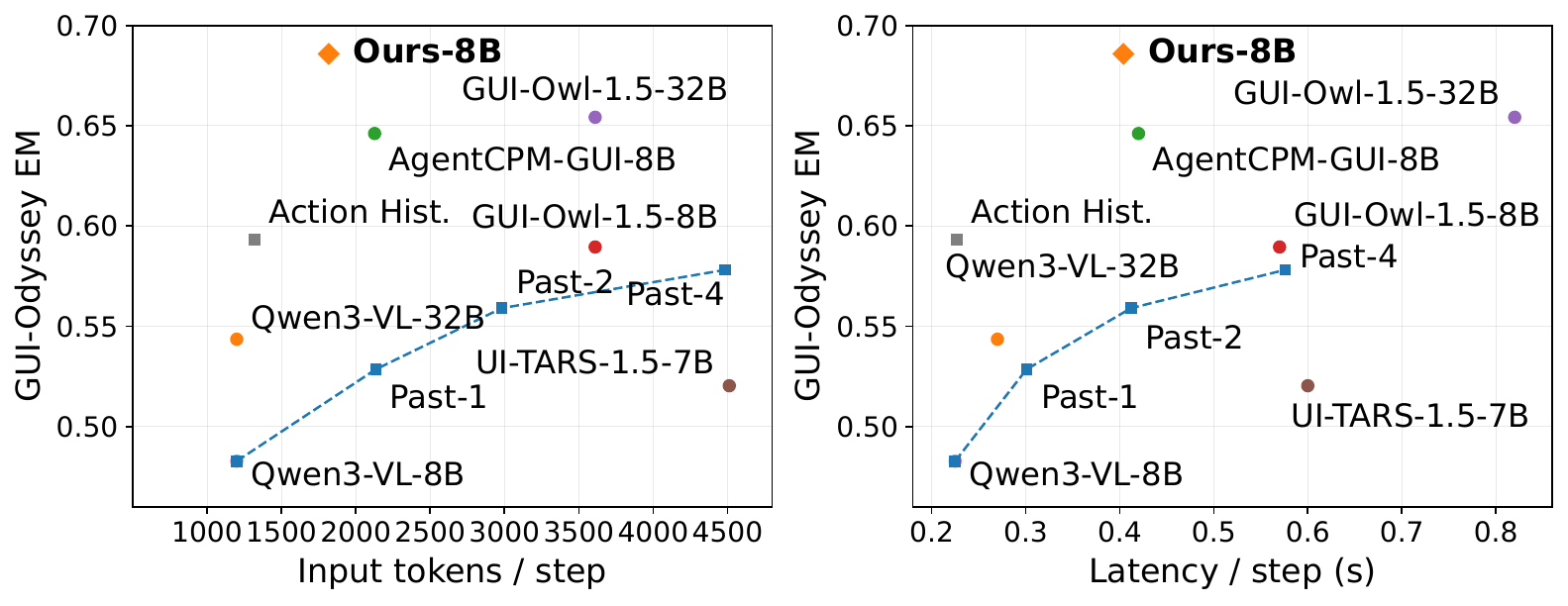}\par
    \vspace{-10pt}

    \caption{\textbf{Efficiency on GUI-Odyssey.} EM versus input tokens and latency; dashed lines connect Qwen3-VL-8B history settings.}
    \label{fig:eff_transfer}
\end{minipage}
\hfill
\begin{minipage}[t]{0.28\linewidth}
    \centering
    \vspace{-4pt}

    \captionof{table}{{Zero-shot transfer on MageBench-Sokoban without additional training.} }
    \label{tab:zero_shot_transfer}

    {\scriptsize
    \setlength{\tabcolsep}{3.5pt}
    \renewcommand{\arraystretch}{1.08}
    \begin{tabular}{lcc}
    \toprule
    Model & Score & Solved \\
    \midrule
    Qwen3-VL-8B & 49.23 & 1/20 \\
    GLM-4.1V-9B & 51.46 & 3/20 \\
    Qwen2.5-VL-72B & 50.74 & 1/20 \\
    UI-TARS-1.5-7B & 51.84 & 2/20 \\
    MAI-UI-8B & 53.03 & 2/20 \\
    \textbf{Ours} & \textbf{55.33} & \textbf{4/20} \\
    \midrule
    $\Delta$ over Base & +6.10 & +3/20 \\
    \bottomrule
    \end{tabular}
    \par}
\end{minipage}

\end{figure}

As shown in Table~\ref{tab:main_results}, ECHOES consistently improves over Qwen3-VL-8B across all four GUI benchmarks, with gains in action prediction, action-type recognition, and visual grounding.
It remains competitive with larger general VLMs and GUI-specialized models while relying only on the current screenshot and recursive $C$, whereas some baselines access ground-truth interaction history.
These results support transferring the decision-making advantage of training-time visual history into a compact deployment representation. Additional results on InternVL3.5-8B support transferability across backbones (Appendix~\ref{app:additional_backbone}).
Beyond GUI tasks, ECHOES also outperforms its base model in normalized task score and fully solved tasks on MageBench-Sokoban under zero-shot evaluation, as shown in Table~\ref{tab:zero_shot_transfer}.
Both models use the same native online interaction pipeline, while ECHOES retains its recursive $C$ state without task-specific adaptation. This suggests that its gains transfer beyond GUI tasks rather than arising from a GUI-specific action or compression format.

Efficiency comparisons further show that ECHOES maintains high performance at relatively low inference cost, as shown in Figure~\ref{fig:eff_transfer}.
Compared with GUI-Owl and UI-TARS under longer-history settings, it uses substantially fewer input tokens and achieves lower per-step latency.
In controlled comparisons with the Qwen3-VL-8B backbone fixed, increasing the number of historical screenshots expands the visual context and raises latency, while performance gains gradually saturate.
Recursive $C$ achieves a better performance--cost trade-off without retaining historical screenshots.
Although action-only history is cheaper, it remains less accurate, suggesting that the benefit comes from modeling task-relevant historical states rather than simply adding textual context. The additional training-time overhead is 16.7\%, as detailed in Appendix~\ref{app:training_overhead}.

\subsection{Ablation Studies}

Table~\ref{tab:ablation}(a) shows that both RL and privileged OPD improve GUI-Odyssey performance from the same SFT initialization. RL optimizes task rewards, while OPD provides dense multi-image supervision on student-generated CTA prefixes. Combining them yields further gains; PMF Gap-guided scheduling outperforms predefined linear OPD--RL scheduling by 3.46 percentage points in EM (68.58 vs.\ 65.12), supporting adaptive coordination beyond a prescribed transition between the two objectives. Trajectory-level paired bootstrap results are provided in Appendix~\ref{app:ablation_significance}.

Inference-time interventions in Table~\ref{tab:ablation}(b) remove memory (No $C$), substitute another trajectory's state (Shuffled $C$), or reuse an outdated state (Stale $C$). All three reduce performance, supporting reliance on trajectory-consistent, updated history rather than additional text alone. Generic recursive summarization (Naive Summary) partially recovers performance but remains below learned recursive updates (Recursive $C$). These results support retaining decision-relevant states rather than merely summarizing the past, consistent with our objective of learning a task-sufficient textual state.

\begin{table}[t]                                                           
\centering
\caption{\textbf{Ablation studies on GUI-Odyssey.}
Left: contributions of RL, privileged OPD, and PMF-guided joint optimization.
Right: interventions on the recurrent textual state $C$.}
\label{tab:ablation}
\renewcommand{\arraystretch}{1.08}

\begin{minipage}[t]{0.55\linewidth}
\centering
\textbf{\small (a) Training objectives}\\[1mm]

\begin{tabular*}{\linewidth}{@{\extracolsep{\fill}}lcc@{}}
\toprule
Variant & TM $\uparrow$ & EM $\uparrow$ \\
\midrule
SFT                                  & 77.00 & 56.24 \\
SFT + RL                             & 79.55 & 63.65 \\
SFT + OPD                            & 81.05 & 64.01 \\
SFT + OPD + RL (Linear $\lambda$)    & 81.31    & 65.12    \\
\textbf{Full (PMF-guided)}           & \textbf{82.25} & \textbf{68.58} \\
\bottomrule
\end{tabular*}

\end{minipage}
\hfill
\begin{minipage}[t]{0.42\linewidth}
\centering
\textbf{\small (b) Compression intervention}\\[1mm]

\begin{tabular*}{\linewidth}{@{\extracolsep{\fill}}lcc@{}}
\toprule
$C$ Setting & TM $\uparrow$ & EM $\uparrow$ \\
\midrule
No $C$                 & 71.72 & 53.39 \\
Shuffled $C$           & 74.29 & 55.21 \\
Stale $C$              & 73.63    & 57.37    \\
Naive Summary          & 77.84 & 60.50 \\
\textbf{Recursive $C$} & \textbf{82.25} & \textbf{68.58} \\
\bottomrule
\end{tabular*}

\end{minipage}
\vspace{-5pt}
\end{table}

\vspace{-5pt}
\subsection{Long-Horizon, Training Dynamics, and Hyperparameter Analysis}

\begin{table}[t]
\centering
\renewcommand{\arraystretch}{1.06}

\begin{minipage}[t]{0.52\linewidth}
\centering
\vspace{0pt}

\captionof{table}{\textbf{Memory dependency.} GUI-Odyssey results on history-dependent (Dep.) and independent (Indep.) steps.}
\label{tab:memory_dependency}

{\footnotesize
\begin{tabular*}{\linewidth}{@{\extracolsep{\fill}}lccc@{}}
\toprule
Model & Overall & Dep. $\uparrow$ & Indep. $\uparrow$ \\
\midrule
Qwen3-VL-32B        & 54.4 & 56.4 & 53.3 \\
UI-TARS-1.5-7B      & 52.0 & 44.0 & 56.5 \\
AgentCPM-GUI-8B     & 64.6 & 59.8 & 67.3 \\
GUI-Owl-1.5-8B      & 59.0 & 58.5 & 59.3 \\
GUI-Owl-1.5-32B     & 65.4 & 65.9 & 65.1 \\
\textbf{Ours-8B}    & \textbf{68.6} & \textbf{72.5} & 66.4 \\
\bottomrule
\end{tabular*}
\par}
\end{minipage}
\hfill
\begin{minipage}[t]{0.44\linewidth}
\centering
\vspace{8pt}

\includegraphics[width=\linewidth]{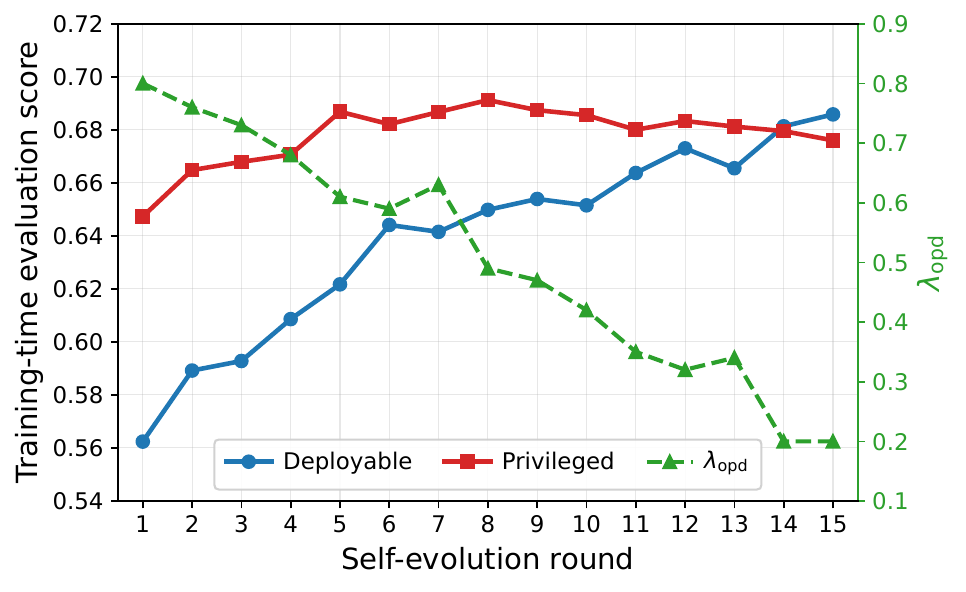}
\vspace{-25pt}
\captionof{figure}{\textbf{Round-wise self-evolution.} Privileged/deployable scores and $\lambda_{\mathrm{OPD}}$.}
\label{fig:self_evolution}
\end{minipage}

\par\vspace{0pt}

\begin{minipage}{\linewidth}
\centering

\captionof{table}{\textbf{Trajectory-length analysis on GUI-Odyssey.} Performance on trajectories with $\leq12$ or $\geq13$ steps, comparing history settings (left) and models (right).}
\label{tab:long_horizon}

\begin{minipage}[t]{0.49\linewidth}
\centering
\vspace{0pt}

\footnotesize
\setlength{\tabcolsep}{4.0pt}
\begin{tabular}{lcc}
\toprule
Model / Setting & $\leq$12 Steps $\uparrow$ & $\geq$13 Steps $\uparrow$ \\
\midrule
Qwen3-VL-8B          & 55.36 & 45.51 \\
Qwen3-VL-8B + Past-1 & 59.47 & 50.27 \\
Qwen3-VL-8B + Past-2 & 62.16 & 53.46 \\
Qwen3-VL-8B + Past-4 & 63.79 & 55.49 \\
Qwen3-VL-8B + Past-all  & 64.54 & 56.29 \\
\textbf{Ours-8B}     & \textbf{70.49} & \textbf{67.84} \\

\bottomrule
\end{tabular}
\end{minipage}
\hfill
\begin{minipage}[t]{0.49\linewidth}
\centering
\vspace{0pt}

\footnotesize
\setlength{\tabcolsep}{4.0pt}
\begin{tabular}{lcc}
\toprule
Model & $\leq$12 Steps $\uparrow$ & $\geq$13 Steps $\uparrow$ \\
\midrule
UI-TARS-1.5-7B      & 59.52 & 49.12 \\
AgentCPM-GUI-8B     & 69.00 & 62.90 \\
GUI-Owl-1.5-8B     & 64.49 & 56.79 \\
GUI-Owl-1.5-32B    & 69.14 & 63.94 \\
Qwen3-VL-32B       & 60.79 & 51.84 \\
\textbf{Ours-8B}   & \textbf{70.49} & \textbf{67.84} \\
\bottomrule
\end{tabular}
\end{minipage}

\end{minipage}
\end{table}

\begin{table}[t]
\centering
\caption{\textbf{Hyperparameter sensitivity on GUI-Odyssey.}
We vary the teacher-refresh interval and the OPD--RL transition range while keeping the remaining training configuration unchanged.}
\label{tab:hyperparameter}
\renewcommand{\arraystretch}{1.08}

\begin{minipage}[t]{0.48\linewidth}
\centering
\textbf{\small (a) Teacher refresh interval}\\[1mm]

\begin{tabular*}{\linewidth}{@{\extracolsep{\fill}}ccc@{}}
\toprule
Steps / Round & TM $\uparrow$ & EM $\uparrow$ \\
\midrule
3             & 81.43 & 67.92 \\
\textbf{6}    & 82.25 & 68.58 \\
12            & 80.79 & 65.84 \\
\bottomrule
\end{tabular*}
\end{minipage}
\hfill
\begin{minipage}[t]{0.48\linewidth}
\centering
\textbf{\small (b) OPD--RL transition range}\\[1mm]

\begin{tabular*}{\linewidth}{@{\extracolsep{\fill}}ccc@{}}
\toprule
$\lambda_{\mathrm{OPD}}$ Range & TM $\uparrow$ & EM $\uparrow$ \\
\midrule
$[0.3,\,0.7]$          & 81.73 & 66.74 \\
$\mathbf{[0.2,\,0.8]}$ & 82.25 & 68.58 \\
$[0.1,\,0.9]$          & 82.08 & 68.21 \\
\bottomrule
\end{tabular*}
\end{minipage}
\vspace{-5pt}
\end{table}

Table~\ref{tab:memory_dependency} shows that ECHOES leads on memory-dependent steps, with larger gains on dependent than independent steps across all listed baselines (e.g., 6.6 vs.\ 1.3 percentage points over GUI-Owl-1.5-32B). It also leads both trajectory-length groups in Table~\ref{tab:long_horizon}, achieving 70.49 EM for $\leq 12$ steps and 67.84 for longer trajectories. Together with the $C$ interventions, these results support the role of updated, trajectory-consistent memory in long-horizon action decision without retaining historical screenshots. Appendix~\ref{app:Case_Study} illustrates history-conditioned decisions on similar screens.

Figure~\ref{fig:self_evolution} shows an overall narrowing of the privileged--deployable test-set gap despite round-wise fluctuations. The deployable score rises from 0.5624 to 0.6858, slightly exceeding the final privileged score of 0.6760. Concurrently, the validation-driven OPD weight decreases overall from 0.8 to 0.2, shifting emphasis toward task-reward optimization. Round-wise test scores and validation-driven weights are reported in Appendix~\ref{app:roundwise_dynamics}.

Under a fixed training budget, Table~\ref{tab:hyperparameter} compares teacher-refresh intervals of 3, 6, and 12 steps and three OPD--RL transition ranges. The default 6-step interval and $\lambda_{\mathrm{OPD}}\in[0.2,0.8]$ achieve the highest EM (68.58) in their respective comparisons; more frequent refreshes or a wider range offer no improvement.

\subsection{Diagnostic Benchmark and PMF Analysis}

Table~\ref{tab:diagnostic_bmk} evaluates element semantics, grounding, whole-screen understanding, and cross-frame state changes.
ECHOES leads all four L2 state-understanding tasks and remains competitive with larger VLMs on Semantic-to-BBox, suggesting improved screen understanding beyond action prediction.
Model rankings vary across perception dimensions, showing that a single grounding or end-to-end metric provides an incomplete assessment.
Stronger direct perception also does not necessarily imply higher memory fidelity.
Using the PMF Gap defined in Sec.~4.1, ECHOES achieves the lowest average gap across the four L2 tasks (0.021 versus 0.054 for Qwen3-VL-8B), alongside higher direct-vision scores, supporting improved preservation of visual states in textual memory.
Task-wise PMF results in Appendix~\ref{app:ECHOESBMK_results} show that Active-Tab generally incurs larger gaps than Foreground and Blocking, including for ECHOES.
Together, these findings highlight improvements in both state understanding and memory fidelity, while identifying fine-grained state retention as a remaining bottleneck not captured by perception accuracy alone.

\begin{table}[t]
\centering
\caption{\textbf{Perception and memory fidelity on ECHOESBMK.} We report direct-vision scores ($\uparrow$) for seven tasks and Avg. PMF ($\downarrow$), the mean nonnegative performance drop from vision to text-only memory replay across four L2 tasks. Bold indicates the best results.}
\label{tab:diagnostic_bmk}
\renewcommand{\arraystretch}{1.06}

\begin{tabular*}{\linewidth}{@{\extracolsep{\fill}}lcccccccc@{}}
\toprule
Model
& B$\rightarrow$F
& S$\rightarrow$B
& Tab
& Fore.
& Block.
& Select.
& Diff
& Avg. PMF $\downarrow$ \\
\midrule

Qwen3-VL-8B      & 0.584 & 0.773 & 0.675 & 0.767 & 0.753 & 0.347 & 0.718 & 0.054 \\
Qwen3-VL-32B     & 0.613 & 0.789 & 0.691 & 0.796 & 0.762 & 0.287 & 0.736 & 0.041 \\
Qwen3.5-27B      & 0.620 & \textbf{0.838} & 0.707 & 0.801 & 0.808 & 0.415 & 0.754 & 0.089 \\
GLM-4.6V-Flash   & 0.582 & 0.752 & 0.634 & 0.792 & 0.848 & 0.307 & \textbf{0.792} & 0.063 \\
Step3-VL-10B     & 0.626 & 0.651 & 0.667 & 0.753 & 0.832 & 0.380 & 0.585 & 0.058 \\
MAI-UI-8B        & 0.569 & 0.804 & 0.650 & 0.723 & 0.826 & 0.318 & 0.633 & 0.059 \\
GUI-Owl-1.5-8B   & 0.538 & 0.813 & 0.675 & 0.814 & 0.814 & 0.370 & 0.729 & 0.067 \\
UI-TARS-1.5-7B   & 0.577 & 0.710 & 0.545 & 0.799 & 0.768 & 0.254 & 0.660 & 0.132 \\
\textbf{Ours-8B} & \textbf{0.673} & 0.821 & \textbf{0.724} & \textbf{0.842} & \textbf{0.870} & \textbf{0.471} & \textbf{0.792} & \textbf{0.021} \\

\bottomrule
\end{tabular*}

\end{table}

\vspace{-5pt}
\section{Conclusion}
\label{sec:conclusion}
\vspace{-5pt}

We present ECHOES, which treats recent visual history as training-time privileged information rather than a persistent deployment input.
It combines CTA recursive state modeling, multi-image--single-image privileged OPD, and PMF Gap-guided OPD--RL with a round-wise refreshed teacher, transferring history-informed decision-making to a policy conditioned on the task instruction, current screenshot, and self-generated textual state $C$.
Experiments show improvements over Qwen3-VL-8B across four GUI benchmarks, favorable performance--cost trade-offs, and zero-shot transfer beyond GUI-specific action spaces.
Memory-dependency analyses, ablations, and Compression interventions support the role of task-relevant historical-state modeling rather than additional text alone.
ECHOESBMK further distinguishes visual-state understanding from memory fidelity, showing improved direct perception and a smaller average PMF Gap, while fine-grained state retention and long-horizon degradation remain challenges.
Together, these findings support using privileged visual history during training to learn compact textual states that preserve decision-relevant information without retaining historical screenshots at deployment.

\section*{AI use statement}

We used generative AI tools to assist with manuscript language refinement, supervised fine-tuning (SFT) data construction, and the construction and verification of ECHOESBMK. All AI-assisted content and data were reviewed by the authors, who take full responsibility for the final manuscript, experimental results, and associated artifacts.

% ...

\bibliography{iclr2027_conference}
\bibliographystyle{iclr2027_conference}

\clearpage

\appendix
\newpage

\section{Additional Experimental Details}
\label{app:experimental_details}

\subsection{Benchmarks and Metrics}

\noindent\textbf{GUI benchmarks.}
We report TM and EM on GUI-Odyssey.
AndroidControl-v2-High reports overall accuracy, action-type accuracy, and grounding accuracy, while AITW and AMEX-L3 report step-level and action-type accuracy.
These metrics characterize complementary aspects of GUI decision-making, including action selection and visual grounding.

\noindent\textbf{Zero-shot visual planning.}
We evaluate MageBench-Sokoban without additional training and report the normalized task score and the number of fully solved tasks.
In the comparison between ECHOES and its Qwen3-VL-8B base model, both models follow the same MageBench native online interaction pipeline.
This evaluation examines whether the training gains transfer beyond GUI tasks.

\noindent\textbf{Diagnostic evaluation.}
ECHOESBMK comprises seven tasks covering element-level perception, whole-screen state understanding, and cross-frame state changes.
In addition to task-specific perception scores, we report the PMF Gap to measure the nonnegative performance drop from direct visual access to reliance on model-generated textual memory.
The reported average PMF Gap is macro-averaged over the four L2 whole-screen state tasks.
Benchmark construction and task-specific evaluation protocols are provided in the diagnostic benchmark appendix.

\subsection{Baseline Inference Context}

We retain favorable model-specific inference settings rather than imposing a uniform current-only input on all baselines.
Official or native inference procedures are used whenever possible, including native or ground-truth interaction history where applicable.
We distinguish four context settings: \emph{current-only}, \emph{native history}, \emph{GT history}, and \emph{Current+$C$}.
Native history follows the baseline's own history-handling mechanism, whereas GT history denotes ground-truth interaction information supplied during evaluation.
In contrast, under Current+$C$, ECHOES receives the task instruction and current screenshot together with its recursively generated Compression, without accessing historical screenshots or ground-truth historical information.
Thus, the cross-model comparison preserves each baseline's supported inference setup rather than enforcing an identical historical-information budget.

\subsection{Efficiency Evaluation}

We jointly report input-token count and wall-clock latency per step to characterize inference cost.
Cross-model comparisons follow the model-specific protocols described above.
Complementarily, controlled comparisons use the same Qwen3-VL-8B backbone with current-only inputs, textual action history, and varying numbers of historical screenshots, alongside the Current+$C$ deployment setting.
These comparisons assess the trade-off between task performance and inference cost under different history representations and context budgets.

\section{Additional Results}
\label{app:results}

\subsection{Evaluation on an Additional VLM Backbone}
\label{app:additional_backbone}

\paragraph{Experimental design.}
To examine whether ECHOES extends beyond Qwen3-VL, we repeat
the training pipeline with InternVL3.5-8B
\citep{wang2025internvl3_5}.
We compare the released checkpoint (Base), CTA supervised
fine-tuning (CTA-SFT), and the full ECHOES model.
The experiments use the same approximately 10K SFT samples
and 9.5K OPD--RL samples, data splits, action space, and
evaluation metrics as the main experiments.
CTA-SFT and ECHOES share the deployment input
$(q,o_t,C_{t-1})$, allowing us to assess the additional
benefit of privileged OPD--RL beyond CTA initialization.
Both the student and the privileged teacher use InternVL3.5-8B.
The teacher receives $(q,o_t,V_t)$ without $C_{t-1}$ and is
refreshed from the student at round boundaries.
The OPD weight follows the same PMF-guided rule but is
computed from InternVL's own validation-set gaps rather
than copied from the Qwen3-VL run.

\paragraph{Results.}
Table~\ref{tab:additional_backbone} compares the two backbones
across four GUI benchmarks.
For InternVL3.5-8B, ECHOES improves over Base by
17.54, 6.45, 9.38, and 10.26 percentage points on
GUI-Odyssey EM, AC-v2-High accuracy, AITW step accuracy,
and AMEX-L3 step accuracy, respectively.
The corresponding gains over CTA-SFT are
8.91, 3.86, 5.24, and 5.44 points.
The gains vary across benchmarks rather than uniformly
matching those of Qwen3-VL.
Our primary comparison is between ECHOES and CTA-SFT
within the same backbone, assessing additional optimization
beyond supervised initialization rather than requiring
one backbone to consistently outperform the other.

\begin{table}[t]
\centering
\caption{
\textbf{Additional-backbone comparison on four GUI benchmarks.}
All scores are percentages, and $\Delta$ denotes differences
in percentage points.
Qwen3-VL-8B results are reproduced from the main-results table.
Base denotes the released checkpoint before our training.
}
\label{tab:additional_backbone}
\small
\setlength{\tabcolsep}{2.4pt}
\renewcommand{\arraystretch}{1.08}

\begin{tabular*}{\linewidth}{@{\extracolsep{\fill}}l*{9}{c}@{}}
\toprule
& \multicolumn{2}{c}{GUI-Odyssey}
& \multicolumn{3}{c}{AC-v2-High}
& \multicolumn{2}{c}{AITW}
& \multicolumn{2}{c}{AMEX-L3} \\
\cmidrule(lr){2-3}
\cmidrule(lr){4-6}
\cmidrule(lr){7-8}
\cmidrule(lr){9-10}
Variant & TM $\uparrow$ & EM $\uparrow$ & Acc $\uparrow$ & Type $\uparrow$ & Ground $\uparrow$ & Step $\uparrow$ & Type $\uparrow$ & Step $\uparrow$ & Type $\uparrow$ \\
\midrule

\multicolumn{10}{l}{\textit{Qwen3-VL-8B}} \\
Base                   & 76.14 & 48.27 & 49.62 & 63.28 & 64.43 & 44.16 & 56.40 & 62.28 & 80.10 \\
ECHOES                 & 82.25 & 68.58 & 53.60 & 76.96 & 65.56 & 59.35 & 66.30 & 73.90 & 84.60 \\
$\Delta$ over Base     & +6.11 & +20.31 & +3.98 & +13.68 & +1.13 & +15.19 & +9.90 & +11.62 & +4.50 \\

\midrule
\multicolumn{10}{l}{\textit{InternVL3.5-8B}} \\
Base                   & 75.43 & 47.19 & 48.67 & 66.42 & 62.57 & 46.83 & 59.74 & 64.56 & 80.27 \\
CTA-SFT                & 77.61 & 55.82 & 51.26 & 71.35 & 64.92 & 50.97 & 63.92 & 69.38 & 82.46 \\
ECHOES                 & 81.07 & 64.73 & 55.12 & 76.28 & 64.76 & 56.21 & 67.53 & 74.82 & 85.14 \\
$\Delta$ over Base     & +5.64 & +17.54 & +6.45 & +9.86 & +2.19 & +9.38 & +7.79 & +10.26 & +4.87 \\

\bottomrule
\end{tabular*}
\end{table}

\subsection{Detailed PMF results on ECHOESBMK}
\label{app:ECHOESBMK_results}

Table~\ref{tab:diagnostic_pmf_detail} breaks down the PMF Gap across the four L2 state categories. ECHOES achieves the lowest gap on Active-Tab, Foreground-Layer, and Screen-Selection, yielding the lowest macro-average of 0.021, compared with 0.054 for Qwen3-VL-8B. Active-Tab incurs the largest gap for every evaluated model; ECHOES reduces this gap from 0.138 to 0.065 relative to its base model, although navigation-state retention remains its largest source of degradation. Foreground-Layer and Screen-Selection are preserved more reliably, with gaps of only 0.000 and 0.008. On Blocking-State, ECHOES ranks second, behind its base model (0.012 versus 0.004). Interpreted alongside the direct-vision scores in Table~\ref{tab:diagnostic_bmk}, these results support improved overall visual-to-text memory fidelity while highlighting fine-grained navigation states as a remaining challenge.

\begin{table}[t]
\centering
\caption{
\textbf{Detailed PMF results on the diagnostic GUI benchmark.}
We report the visual-to-text performance gap for each of the four L2 state categories.
Lower PMF indicates less information loss during visual-to-text transformation.
}
\label{tab:diagnostic_pmf_detail}
\renewcommand{\arraystretch}{1.06}

\begin{tabular*}{\linewidth}{@{\extracolsep{\fill}}lccccc@{}}
\toprule
Model
& Tab PMF $\downarrow$
& Fore. PMF $\downarrow$
& Block. PMF $\downarrow$
& Select. PMF $\downarrow$
& Avg. PMF $\downarrow$ \\
\midrule

Qwen3-VL-8B      & 0.138 & 0.050 & \textbf{0.004} & 0.025 & 0.054 \\
Qwen3-VL-32B     & 0.089 & 0.040 & 0.016 & \underline{0.017} & \underline{0.041} \\
Qwen3.5-27B      & 0.179 & 0.076 & 0.077 & 0.025 & 0.089 \\
GLM-4.6V-Flash   & \underline{0.073} & 0.059 & 0.053 & 0.068 & 0.063 \\
Step3-VL-10B     & 0.106 & 0.019 & 0.065 & 0.042 & 0.058 \\
MAI-UI-8B        & 0.114 & 0.049 & 0.049 & 0.025 & 0.059 \\
GUI-Owl-1.5-8B   & 0.154 & \underline{0.003} & 0.033 & 0.076 & 0.067 \\
UI-TARS-1.5-7B   & 0.390 & 0.014 & 0.049 & 0.076 & 0.132 \\
\textbf{Ours-8B} & \textbf{0.065} & \textbf{0.000} & \underline{0.012} & \textbf{0.008} & \textbf{0.021} \\

\bottomrule
\end{tabular*}

\end{table}

\subsection{Round-wise Training Dynamics}
\label{app:roundwise_dynamics}

Table~\ref{tab:roundwise_schedule} reports GUI-Odyssey test-set EM for the privileged and deployable views of the same model checkpoint, together with the logged OPD weight. Importantly, $S_r^P$ and $S_r^D$ are measured on the test set, whereas $\lambda_r$ is determined from the privileged--deployable gap on a separate validation set at round boundaries; the displayed test-set gap is not used to compute the weight. The deployable score increases from 0.5624 to 0.6858, while the signed test-set gap narrows from +0.0849 to -0.0098, with the final deployable view slightly outperforming the privileged view. Meanwhile, $\lambda_r$ decreases from 0.80 to 0.20, showing that the narrowing test-set gap accompanies the validation-driven shift toward greater emphasis on task-reward optimization.

\begin{table}[t]
\centering
\caption{\textbf{Round-wise privileged--deployable gap and OPD weight.}}
\label{tab:roundwise_schedule}
\renewcommand{\arraystretch}{1.06}

\begin{tabular*}{\linewidth}{@{\extracolsep{\fill}}ccccc@{}}
\toprule
Round
& $S^P_r$
& $S^D_r$
& $S^P_r-S^D_r$
& $\lambda_r$ \\
\midrule
0     & 0.6473 & 0.5624 & +0.0849 & 0.80 \\
3     & 0.6706 & 0.6086 & +0.0620 & 0.68 \\
6     & 0.6867 & 0.6415 & +0.0452 & 0.63 \\
9     & 0.6855 & 0.6515 & +0.0340 & 0.42 \\
12    & 0.6812 & 0.6655 & +0.0157 & 0.34 \\
Final & 0.6760 & 0.6858 & -0.0098 & 0.20 \\
\bottomrule
\end{tabular*}
\end{table}

\subsection{Statistical Significance of Ablation Results}
\label{app:ablation_significance}

Table~\ref{tab:ablation_significance} reports paired EM improvements of the full ECHOES model over its training ablations on GUI-Odyssey, with 95\% confidence intervals obtained using trajectory-level paired bootstrap with 10,000 resamples. ECHOES improves over SFT, SFT+RL, and SFT+OPD by 12.34, 4.93, and 4.57 percentage points, respectively. All reported confidence intervals exclude zero, and all reported $p$-values are at most 0.002. In particular, the improvement over linear-weight OPD--RL is 3.46 percentage points, with a 95\% confidence interval of $[1.77,5.12]$ and $p=0.002$, supporting the additional contribution of PMF Gap-guided scheduling beyond combining the two training objectives with a linear weight.

\begin{table}[t]
\centering
\caption{
\textbf{Statistical significance of the ablation comparisons on GUI-Odyssey.}
We report paired EM improvements of the full model over each training
variant, with 95\% confidence intervals obtained using trajectory-level
paired bootstrap with 10,000 resamples.
}
\label{tab:ablation_significance}
\renewcommand{\arraystretch}{1.06}

\begin{tabular*}{\linewidth}{@{\extracolsep{\fill}}lccc@{}}
\toprule
Comparison
& $\Delta$ EM $\uparrow$
& 95\% CI
& $p$-value \\
\midrule
Full vs. SFT
& +12.34 & [10.61, 14.07] & $<0.001$ \\

Full vs. SFT + RL
& +4.93 & [3.24, 6.60] & $<0.001$ \\

Full vs. SFT + OPD
& +4.57 & [2.90, 6.22] & $<0.001$ \\

Full vs. OPD + RL (Linear $\lambda$)
& +3.46 & [1.77, 5.12] & 0.002 \\
\bottomrule
\end{tabular*}
\end{table}

\subsection{Training-Time Overhead}
\label{app:training_overhead}

We characterize the additional training-time cost of ECHOES relative to standard OPD--RL using normalized wall-clock time, with the baseline runtime set to 100. The reported 16.7\% overhead corresponds to a normalized ECHOES runtime of 116.70. Table~\ref{tab:training_overhead} provides a training-time overhead result across round-boundary validation and teacher-refresh operations, based on an unchanged baseline training workload and non-overlapping additional operations.

\begin{table}[t]
\centering
\caption{\textbf{Training-time overhead.} Runtime is normalized to the OPD--RL baseline.}
\label{tab:training_overhead}
\small
\setlength{\tabcolsep}{6pt}
\renewcommand{\arraystretch}{1.12}
\begin{tabular*}{\linewidth}{@{\extracolsep{\fill}}lrrr@{}}
\toprule
Component & OPD--RL & ECHOES & Additional time \\
\midrule
Baseline training workload & 100.00 & 100.00 & 0.00 \\
Additional deployable-view validation & 0.00 & 4.83 & 4.83 \\
Additional privileged-view validation & 0.00 & 9.12 & 9.12 \\
Additional teacher refresh/synchronization & 0.00 & 2.75 & 2.75 \\
\midrule
Total normalized runtime & 100.00 & 116.70 & 16.70 \\
Relative overhead & 0.0\% & 16.7\% & --- \\
\bottomrule
\end{tabular*}
\end{table}

The relative overhead is computed as
\[
\Delta_{\mathrm{time}}
=
\frac{\tau_{\mathrm{ECHOES}}-\tau_{\mathrm{OPD\text{-}RL}}}
{\tau_{\mathrm{OPD\text{-}RL}}}
\times 100\%
=
16.7\%.
\]
Under the illustrative allocation, dual-view validation contributes 13.95 normalized time units, while teacher refresh and synchronization contribute 2.75. Validation estimates the privileged--deployable gap for updating $\lambda_{\mathrm{OPD}}$, whereas teacher refresh transfers the latest student parameters to the frozen teacher used in the subsequent round. Any corresponding operations already performed by the baseline belong to the baseline workload rather than the additional-cost terms.

These operations are confined to training. Deployment retains the task instruction, current screenshot, and recursively generated state $C$, requiring neither dual-view validation nor teacher refreshes. The additional training-time cost therefore does not introduce corresponding operations into the inference pipeline.

\section{Case Study}
\label{app:Case_Study}

\begin{figure}[t]
    \centering
    \includegraphics[width=\linewidth]{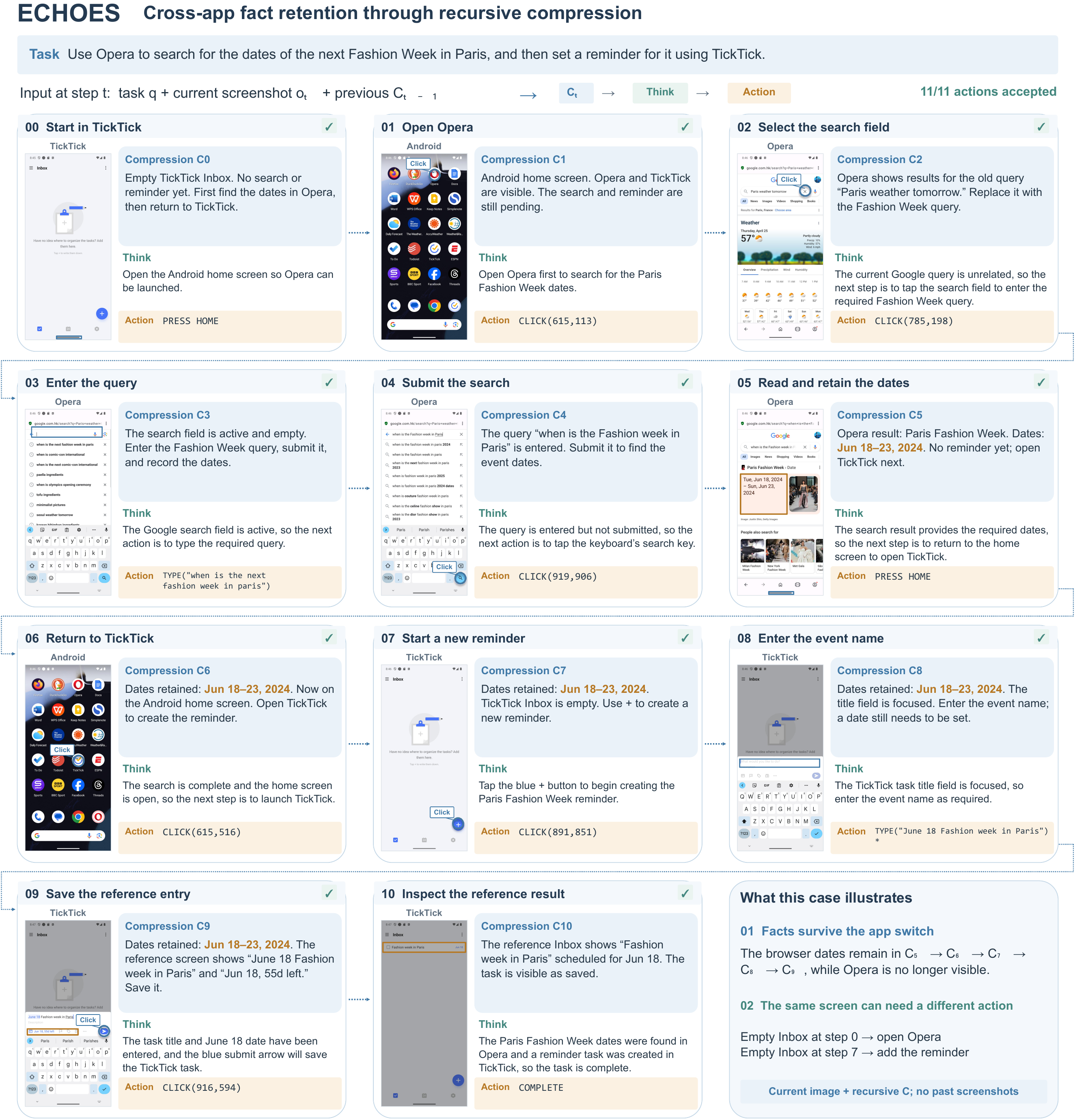}
    \caption{
    \textbf{Cross-app fact retention through recursive compression.}
    An 11-step example involving information retrieval in Opera
    and reminder creation in TickTick.
    Each panel shows the current screenshot and the corresponding
    Compression, Think, and Action outputs.
    Dates recorded at step 05 remain available after leaving Opera,
    while similar Inbox screens at steps 00 and 07 correspond to
    different task stages and actions.
    Each decision uses only the task instruction, current screenshot,
    and previous compression, without historical screenshot inputs.
    }
    \label{fig:case_study}
\end{figure}

Figure~\ref{fig:case_study} presents an 11-step cross-app example
(steps 00--10) involving a search for Paris Fashion Week dates
in Opera and reminder creation in TickTick.
At each step, ECHOES conditions on the task instruction $q$,
current screenshot $o_t$, and previous compression $C_{t-1}$
to generate $(C_t,T_t,A_t)$.
The example highlights two roles of recursive compression:
preserving facts across application switches and tracking
task progress to distinguish decisions on visually similar screens.

\paragraph{Cross-app fact retention.}
At step 05, $C_5$ records the date range June 18--23, 2024,
from the displayed search result and notes that the reminder
has not yet been created.
These dates remain in $C_6,\ldots,C_9$ as the current observation
changes from Opera to the Android home screen and then TickTick.
Meanwhile, the compression updates the immediate task state,
including whether the reminder interface is open and whether
the title field is focused.
The later reference screenshots show a June 18 reminder entry
and its saved state, which are reflected in $C_9$ and $C_{10}$.
Thus, the textual state retains task-relevant facts while
incorporating newly observed progress, without requiring
the earlier browser screenshot as an input.

\paragraph{History-conditioned action selection.}
Steps 00 and 07 show similar empty TickTick Inbox screens,
but correspond to different task stages.
At step 00, the event dates have not been retrieved, so the
model selects \texttt{PRESS\_HOME} to begin navigating to Opera.
At step 07, the search is complete and the dates are retained
in the compression, so the model instead selects the add button
to begin creating the reminder.
This contrast illustrates why current-screen appearance alone
does not specify the appropriate next action:
the decision also depends on what has already been accomplished.
Recursive compression represents this distinction through
both retained facts and updated task progress.

\section{Training and Implementation Details}
\label{app:training_details}

\noindent\textbf{Training data and supervision.}
We use Qwen3-VL-8B as the base model and construct all post-training data
ourselves. The training set contains approximately 10K SFT samples and
9.5K OPD--RL samples. Each SFT sample follows the
Compression--Think--Action (CTA) format and includes explicit supervision
for the recurrent textual state $C$. The target $C$ annotations are generated
by GPT-5.6 using the task instruction and the corresponding interaction
trajectory as context, providing an initialization for task-state writing
before on-policy optimization. The training and evaluation data are constructed
from disjoint splits: no evaluation sample or trajectory is included in either
SFT or OPD--RL training.

\noindent\textbf{Deployable and privileged observations.}
At interaction step $t$, the deployable policy receives only the task
instruction $q$, the current screenshot $o_t$, and the recursively generated
state $C_{t-1}$:
\[
x_t^{D}=(q,o_t,C_{t-1}).
\]
The privileged teacher receives up to four preceding screenshots.
Accordingly, its visual input contains at most five temporally ordered
observations, consisting of the current screenshot and the four most recent
historical screenshots:
\[
V_t=(o_{\max(1,t-4)},\ldots,o_{t-1}),
\qquad
x_t^{P}=(q,o_t,\mathcal{V}_t).
\]
When fewer than four historical observations are available, the privileged
view uses the available history. The teacher and student share the same model architecture, tokenizer, and CTA output format, but receive different observation contexts: $(q,o_t,V_t)$ and $(q,o_t,C_{t-1})$, respectively. The teacher is frozen within each training round and refreshed from the latest student at the round boundary. Compression is constrained to a maximum length of $B_C=128$ tokens.

\noindent\textbf{OPD--RL optimization.}
During OPD--RL training, the deployable policy first generates CTA rollouts
under $x_t^{D}$. The privileged teacher does not generate an independent
trajectory; instead, it performs teacher-forced scoring on the same
student-generated CTA sequence. OPD is applied to valid tokens in
Compression, Think, and Action, while GRPO performs group-relative
optimization over the student rollouts.

The GRPO reward combines action correctness, output-format validity, and a
length penalty on Compression:
\[
R =
R_{\mathrm{act}}
+
R_{\mathrm{fmt}}
-
R_{\mathrm{len}}.
\]
Here, $R_{\mathrm{act}}$ evaluates whether the predicted action type and its
associated grounding or parameters match the target; $R_{\mathrm{fmt}}$
encourages valid and parseable CTA outputs; and $R_{\mathrm{len}}$ discourages
unnecessarily verbose $C$. Unless otherwise specified, the teacher is
refreshed every 6 optimization steps. The default PMF-guided OPD--RL
transition range is
$\lambda_{\mathrm{OPD}}\in[0.2,0.8]$, with the weight updated only at
round boundaries according to the privileged--deployable performance gap.
For GRPO, we use a rollout group size of $K=8$ and sample trajectories with
temperature $1.0$.

\noindent\textbf{Unified action space.}
We normalize the GUI post-training trajectories into the following nine
basic action categories.

\begin{table}[h]
\centering
\caption{\textbf{Unified GUI action space used for post-training.}}
\label{tab:appendix_action_space}
\renewcommand{\arraystretch}{1.05}
\begin{tabular*}{\linewidth}{@{\extracolsep{\fill}}lll@{}}
\toprule
Action & Parameter & Description \\
\midrule
CLICK          & $(x,y)$            & Click a target position \\
LONG\_PRESS    & $(x,y)$            & Long-press a target position \\
TYPE           & text               & Enter textual content \\
SCROLL         & direction          & Scroll up, down, left, or right \\
PRESS\_BACK    & --                 & Return to the previous screen \\
PRESS\_HOME    & --                 & Return to the home screen \\
PRESS\_RECENT  & --                 & Open the recent-apps interface \\
COMPLETE       & --                 & Terminate with task completion \\
IMPOSSIBLE     & --                 & Terminate when the task is infeasible \\
\bottomrule
\end{tabular*}
\end{table}

\section{Related Work}
\label{app:total_related_work}

\subsection{Long-Horizon VLM Agent Reasoning}

With the development of multimodal large language models, visual Agents have gradually expanded from single-step interface understanding and grounding to long-horizon tasks involving planning, reasoning, and sequential decision-making. Early works such as SeeClick, Aguvis, ShowUI, and UI-TARS mainly improve GUI operation capabilities by enhancing visual grounding, action modeling, and explicit reasoning~\citep{cheng2024seeclick,xu2025aguvis,lin2025showui,qin2025uitars}; however, as interaction trajectories grow, current decisions increasingly depend on previous interface states, action outcomes, and task progress, making historical information an important component of long-horizon reasoning. Recent works further focus on history modeling; for example, GUI-Rise, HiconAgent, PAL-UI, AgentProg, and Mem-W retain long-term interaction information through textual summarization, history compression, visual retrieval, structured states, or latent-space memory, respectively~\citep{liu2025guirise,zhou2026hiconagent,liu2025palui,tian2025agentprog,zhang2026memw}. These methods demonstrate the importance of historical context for long-horizon visual decision-making, but generally still rely on the continuous storage, compression, or access of historical information during inference, and therefore require a trade-off between information completeness and visual tokens, context length, and computational overhead. Different from this inference-time history management paradigm, this work explores how to use the recent visual history only as training-time privileged information and transfer its decision-making advantage to a deployment mode that relies only on the current observation and a compact textual state.

\subsection{Agentic Post-Training and Self-Evolution}

Beyond context modeling, post-training paradigms have also continued to advance the long-horizon capabilities of VLM Agents. Traditional supervised fine-tuning relies on fixed expert trajectories and has difficulty covering the state distribution induced by the model's own execution~\citep{ross2011dagger}; reinforcement learning represented by GRPO~\citep{shao2024deepseekmath} further optimizes reasoning and action capabilities through model-generated rollouts and task rewards, and has been applied to GUI Agents by UI-R1, GUI-R1, AgentCPM-GUI, GUI-Rise, UI-TARS-2, and other works~\citep{lu2026uir1,luo2025guir1,zhang2025agentcpm,liu2025guirise,wang2025uitars2}. However, outcome rewards in long-horizon environments are usually relatively sparse, making it difficult for intermediate decisions to receive fine-grained supervision~\citep{lin2026uivoyager,zhang2026guisd}. Complementarily, GKD/On-Policy Distillation (OPD) uses teacher distributions to provide dense token-level signals on states visited by the student itself~\citep{agarwal2024onpolicy}; recent works including Self-Distilled Reasoner, GUI-SD, LiteGUI, and GHD have further explored same-model self-distillation with privileged contexts and the combination of OPD with RL~\citep{zhao2026selfdistilled,zhang2026guisd,wu2026litegui,li2026ghd}. Meanwhile, UI-Voyager and SGCD construct new supervision by continuously utilizing trajectories generated by the model itself, reflecting the trend of Agent training from static data toward self-evolution~\citep{lin2026uivoyager,fan2026sgcd}. Existing works mainly use answers, target regions, future states, or external feedback to improve the current policy, whereas this work focuses on the cross-observation-mode capability gap caused by visual-history deployment constraints, and further dynamically coordinates privileged-policy distillation and task-reward optimization according to this gap, allowing the model to gradually transition from imitating a stronger observation mode to directly optimizing the actual task objective.

\section{Diagnostic Benchmark Details}
\label{app:benchmark}

\subsection{Task Construction and Statistics}
\label{app:benchmark_construction}

\noindent\textbf{Data construction.} ECHOESBMK is grounded in manually designed and reviewed tasks involving real applications. We execute these tasks with multiple GUI agents and derive diagnostic queries from their observations and failure records. Screenshots, component annotations, and interaction trajectories are standardized into records containing image identifiers, dimensions, element annotations, and task-specific reference answers. Cross-frame records additionally link temporally ordered observations to the corresponding interaction context. The benchmark preserves task-native outputs rather than converting all queries into action prediction.

\noindent\textbf{Tasks and statistics.} The benchmark contains 9,061 instances across seven tasks. At L1, \emph{BBox-to-Function} (4,102) identifies the function of a localized element, while \emph{Semantic-to-BBox} (604) localizes a control from a functional request. At L2, \emph{Active-Tab} (123) identifies the active navigation item; \emph{Foreground-Layer} (577) detects and describes foreground interface layers; \emph{Blocking-State} (246) determines whether the requested interaction is blocked; and \emph{Screen-Selection} (118) identifies selected items and values. At L3, \emph{Screen-State Diff} (3,291) describes changes between pre- and post-action screenshots, including navigation, overlay, and selection-state transitions. Multiple queries may share a screenshot, while cross-frame instances contain two observations; instance counts therefore differ from unique-image counts. Alternative MCQ, judge, and memory-replay views are not counted as additional instances.

\subsection{Annotation and Quality Control}
\label{app:benchmark_quality}

\noindent\textbf{Review procedure.} Following multi-model execution and candidate construction, GPT-5.6 Sol~\citep{openai2026gpt56} and Claude Opus 5~\citep{anthropic2026claudeopus5} perform preliminary checking and filtering, followed by sample-by-sample human verification. Reviewers check screenshot--question consistency, target-element and state annotations, and whether the reference answer is unambiguously supported by the supplied observations. Element-level checks verify bounding boxes and functional descriptions; screen-level checks verify navigation, foreground, blocking, and selection states; cross-frame checks verify the observed transition and its direction. Ambiguous or incorrectly annotated instances are removed.

\noindent\textbf{Verification coverage.} All 9,061 retained instances receive human review, covering 4,706 L1, 1,064 L2, and 3,291 L3 instances. This constitutes 100\% review coverage, not an independently measured annotation-accuracy rate. Final annotations are determined through human inspection of visual evidence rather than model agreement alone. Review distinguishes foreground presence from interaction blocking, element visibility from selection, and intended actions from observed outcomes. Benchmark reference annotations are verified separately from the model-generated summaries evaluated for memory fidelity.

\subsection{Evaluation Protocols}
\label{app:benchmark_evaluation}

\noindent\textbf{Task-specific scoring.} Semantic-to-BBox uses region hit rate: predicted coordinates are mapped to the original image and checked against the reference bounding box, with parsing failures recorded separately. Active-Tab, Blocking-State, and foreground-presence judgments use accuracy, while Screen-Selection uses rule-based matching of selected items and values. BBox-to-Function and Screen-State Diff support MCQ and LLM-Judge~\citep{zheng2023judging} evaluation; foreground descriptions can additionally be assessed semantically. MCQ distractors are preferentially drawn from same-screen elements or semantically similar alternatives. Task-specific answer extraction preserves correctness-critical distinctions, including numerical values, state polarity, element identity, and transition direction.

\noindent\textbf{Direct vision and memory replay.} The 1,064 L2 instances support paired evaluation. Each model first generates its own textual screen summary, covering foreground information, the active navigation tab, and selected items or values. Summaries are cached for reuse across applicable queries. The same model then answers each query using either the screenshot or its generated summary alone, with identical questions and reference answers across conditions. Memory-only inputs contain no screenshot and do not substitute human reference captions or summaries from another model. This screen-level protocol measures visual-to-text state retention, complementing the long-horizon experiments on recursive CTA execution.

\noindent\textbf{PMF Gap.} For L2 task $k$, let $S_k^V$ and $S_k^M$ denote normalized task scores on matched instances under direct-vision and memory-only inputs, respectively. We compute
\[
G_{\mathrm{PMF},k}=\max(0,S_k^V-S_k^M), \qquad \overline{G}_{\mathrm{PMF}}=\frac{1}{4}\sum_{k\in\mathcal{K}_{L2}}G_{\mathrm{PMF},k},
\]
where $\mathcal{K}_{L2}$ contains the four whole-screen tasks. The nonnegative operation is applied to each task-level difference before unweighted macro-averaging; L1 and L3 are excluded. Smaller gaps indicate less performance degradation after replacing visual evidence with text, but must be interpreted alongside direct-vision scores rather than as proof of lossless memory. These diagnostic test scores are separate from the validation-set privileged--deployable gap used to schedule OPD--RL training.

\subsection{Task Examples in ECHOESBMK}
\label{app:echoesbmk_examples}

\begin{figure}[t]
    \centering
    \includegraphics[width=\linewidth]{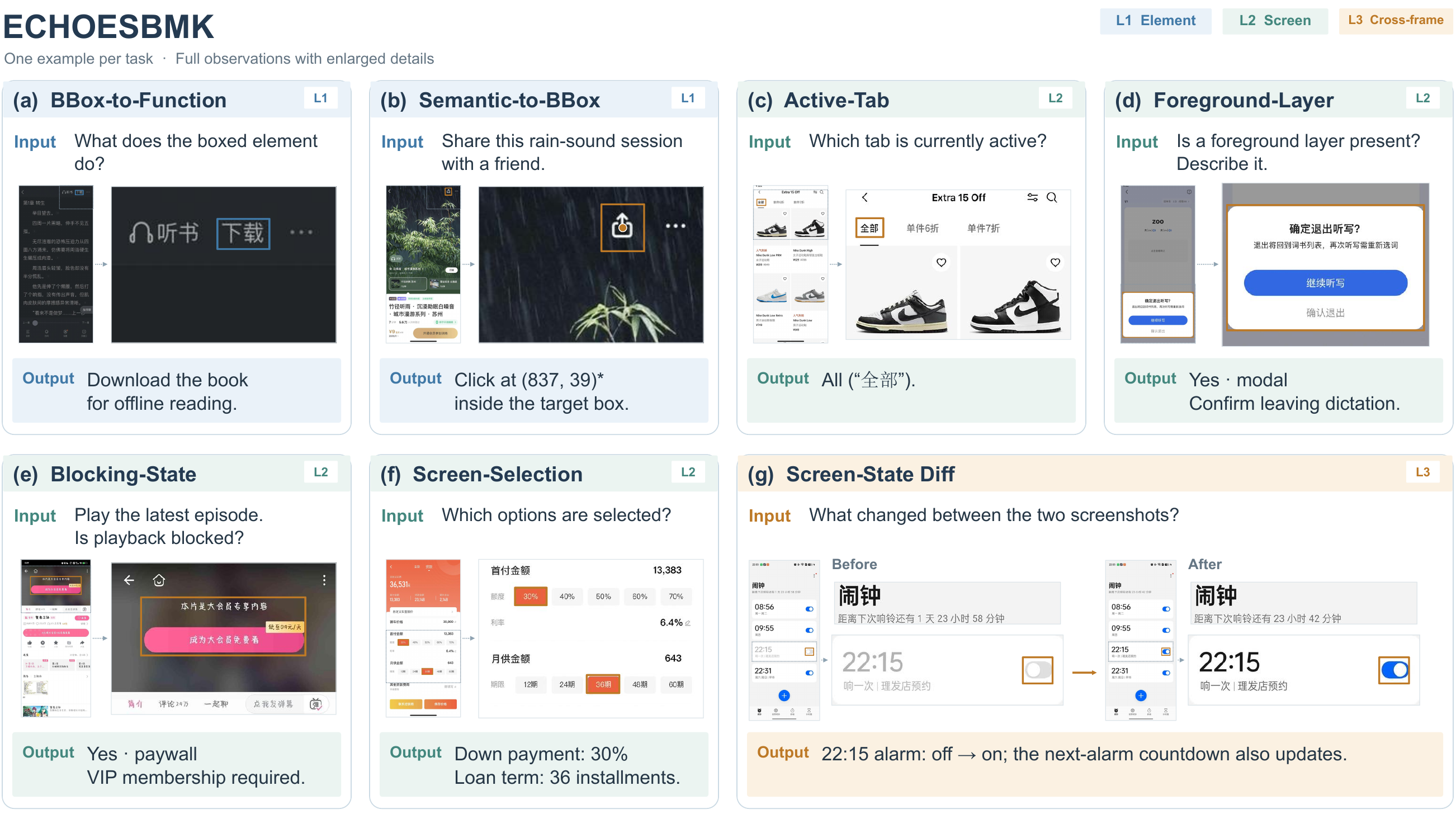}
    \caption{
    \textbf{Representative examples of the seven ECHOESBMK tasks.}
    Panels (a)--(b) illustrate L1 element-level perception:
    BBox-to-Function and Semantic-to-BBox.
    Panels (c)--(f) illustrate L2 whole-screen state understanding:
    Active-Tab, Foreground-Layer, Blocking-State, and Screen-Selection.
    Panel (g) illustrates L3 cross-frame state changes through
    Screen-State Diff.
    Each example includes a task prompt, visual observation(s),
    and a corresponding answer.
    Full screenshots are accompanied by enlarged details
    for readability.
    }
    \label{fig:echoesbmk_examples}
\end{figure}

Figure~\ref{fig:echoesbmk_examples} presents one example for each of the seven diagnostic tasks in ECHOESBMK. The tasks span three levels: element-level perception (L1), whole-screen state understanding (L2), and cross-frame state changes (L3). Each panel pairs a task prompt with the corresponding visual observation and example answer, with enlarged regions highlighting the relevant interface details.

\paragraph{L1: Element-level perception.}
Panels (a) and (b) illustrate two complementary directions
of element understanding.
\textit{BBox-to-Function} asks the model to infer the function
of a highlighted interface element, such as recognizing that
the boxed download button saves a book for offline reading.
Conversely, \textit{Semantic-to-BBox} provides a functional
description and asks the model to localize the corresponding
element, such as identifying the share button for a rain-sound
session.
Together, these tasks examine both the interpretation of
localized visual elements and the grounding of semantic
requests in the interface.

\paragraph{L2: Whole-screen state understanding.}
Panels (c)--(f) examine four aspects of the current screen state.
\textit{Active-Tab} identifies the selected navigation tab,
which is ``All'' in the example.
\textit{Foreground-Layer} detects and describes a foreground
interface layer, here a confirmation modal for leaving dictation.
\textit{Blocking-State} determines whether the requested
interaction is blocked and identifies the blocking condition;
the example shows a VIP-membership paywall preventing playback.
Unlike foreground-layer detection, this task concerns whether
the requested operation can proceed, rather than merely whether
an overlay is present.
Finally, \textit{Screen-Selection} identifies currently selected
options, illustrated by a 30\% down payment and a loan term
of 36 installments.
These tasks assess navigation state, foreground presentation,
interaction constraints, and selected values beyond locating
individual interface elements.

\paragraph{L3: Cross-frame state changes.}
Panel (g) illustrates \textit{Screen-State Diff}, which compares
a before--after screenshot pair rather than a single observation.
The example requires identifying that the 22:15 alarm changes
from off to on and that the next-alarm countdown also updates.
The target is the change between observations, rather than
an independent description of each screen.

The examples retain task-specific output formats, including
functional descriptions, coordinate localization, categorical
judgments, selected-value sets, and change descriptions,
rather than uniformly converting all tasks into action prediction.
They illustrate the direct-vision component of ECHOESBMK;
memory-fidelity analysis additionally compares direct-vision
and text-only memory-replay scores on the four L2 tasks.

\end{document}